\documentclass[letterpaper, 10 pt, conference]{ieeeconf}  

\IEEEoverridecommandlockouts                              

\usepackage{amsmath} 
\usepackage{graphicx}
\usepackage{subfig}
\usepackage{changepage}
\usepackage[mathletters]{ucs}
\usepackage[utf8x]{inputenc}
\usepackage{bm}
\usepackage{xcolor}
\usepackage{cite}
\usepackage[normalem]{ulem}
\usepackage{float}
\usepackage{color}
\usepackage{balance}

\title{\LARGE \bf
CovarianceNet: Conditional Generative Model for Correct Covariance Prediction in Human Motion Prediction
}

\author{Aleksey Postnikov$^{1,2}$ \and Aleksander Gamayunov$^{1,2}$ \and Gonzalo Ferrer$^{2}$%
        \thanks{\textsuperscript{1} The authors are with the Sberbank Robotics Laboratory, Moscow, Russia.
                    {\tt\small \{postnikov.a.l,gamayunov.a.r\}@sberbank.ru}.
               }
        \thanks{\textsuperscript{2}Skolkovo Institute of Science and Technology, Moscow, Russia.
                    {\tt\small g.ferrer@skoltech.ru}.
           }    
    }

\begin{document}

\maketitle
\thispagestyle{empty}
\pagestyle{empty}


\begin{abstract}
The correct characterization of uncertainty when predicting human motion is equally important as the accuracy of this prediction.
We present a new method to correctly predict the uncertainty associated with the predicted distribution of future trajectories. Our approach, CovariaceNet, is based on a Conditional Generative Model with  Gaussian latent variables in order to predict the parameters of a bi-variate Gaussian distribution.
The combination of CovarianceNet with a motion prediction model results in a hybrid approach that outputs a uni-modal distribution.
We will show how some state of the art methods in motion prediction become overconfident when predicting uncertainty, according to our proposed metric and validated in the ETH data-set \cite{pellegrini2009you}.
CovarianceNet correctly predicts uncertainty, which makes our method suitable for applications that use predicted distributions, e.g., planning or decision making.

\end{abstract}
\section{Introduction}


Human Motion Prediction, during the last years,  has received the attention of the research community from different fields: intelligent vehicles, pattern recognition, graphics, robotics, etc.
The motivation to understand and predict human motion is immense and it has a deep impact in related topics, such as, decision making, path planning, autonomous navigation, surveillance, tracking, scene understanding, anomaly detection, etc.

The problem of forecasting where pedestrians will be in the near future is, however, ill-posed by nature: human beings tend to be unpredictable on their decisions and motion is neither exempt of it.
These random nature in motion brings an open challenge to prediction algorithms, where algorithms are desired to be accurate and correctly grasp the uncertainty associated with their predictions.

\begin{figure}[t]
    \center{\includegraphics[width=.34\textwidth]{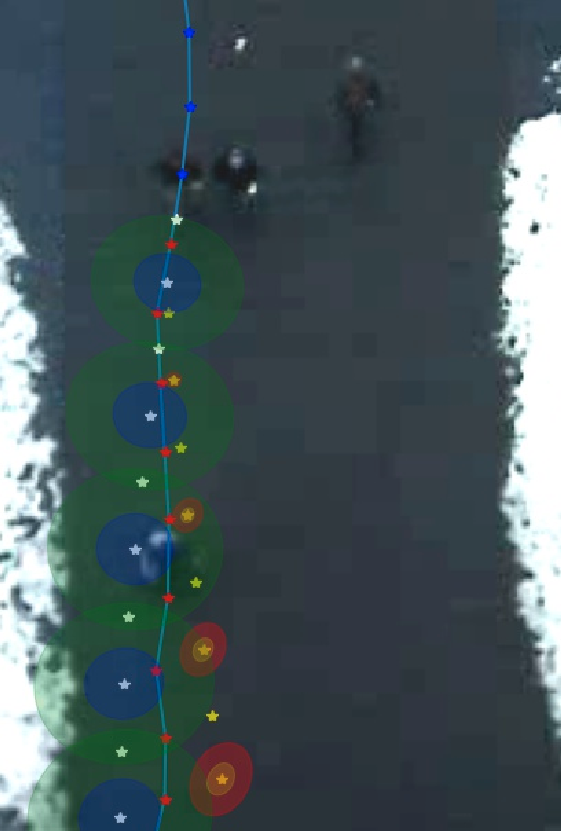}}
    \caption{Comparison of two predicted trajectories in the ETH dataset\cite{pellegrini2009you}. CovarianceNet in cyan stars, blue and green ellipses, - mean, 1 sigma, 3 sigma, respectively. Trajectron++  \cite{salzmann2020trajectron} in yellow stars, yellow and red ellipses, red ellipses - mean, 1 sigma, 3 sigma, respectively. The environment is challenging, since there is a pedestrian in the way.}
    \label{fig:example}
\end{figure}

To this end, multiple benchmarks have been created and released \cite{pellegrini2009you,leal2014learning,caesar2020nuscenes}, providing common grounds to test and evaluate.
Most modern motion prediction algorithms focus on accurate prediction of agent position errors on these benchmarks. Nonetheless, the precision due to this inherent uncertainty is equally important, and this paper is an effort to research on this direction. Prediction algorithms should address this issue as well: it provides a high degree of interpretability by estimating the associated uncertainty and it might be of some use for consequent algorithms making use of prediction information, e.g., planning.
For example, Fig.~\ref{fig:example} shows two predictions, of similar predicted error, but different uncertainty estimation.

\begin{figure}[h]
    \includegraphics[width=.48\textwidth]{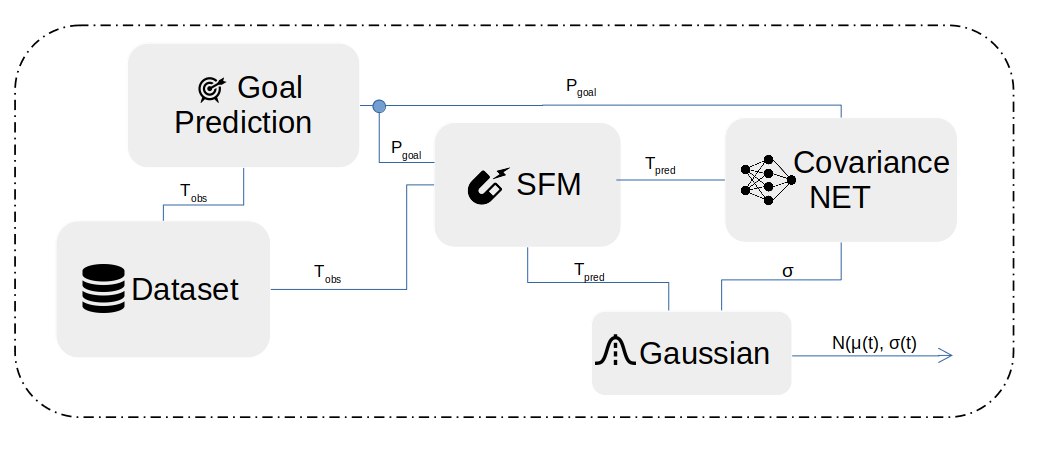}
    \caption{  Our Hybrid approach for Human Motion Prediction and Uncertainty Estimation. We combine customized deep goal prediction model\cite{giuliari2020transformer}  with SFM\cite{helbing1995social}, for mean pedestrian pose prediction and Covariance Net uses Conditional Variational Autoencoder\cite{sohn2015learning}  deep generative model for uncertainty prediction. }
    \label{fig::architecture}
\end{figure}

In this paper, we focus on predicting uncertainty and human motion prediction. We propose a hybrid approach shown on Fig. \ref{fig::architecture}, consisting of  deep model for Goals Prediction and model-based trajectory prediction, complemented by a modern neural-net approach to predict motion as a uni-modal Gaussian distribution.
Several works provide a multi-modal distribution for motion prediction such as \cite{salzmann2020trajectron}, achieving excellent results in benchmarks. However, combining a mixture distribution into robot planning approaches (for instance) requires careful considerations.
Sampling based techniques require extra attention when considering multi-modality since low-probability prediction outcomes might result in dangerous outcomes, when evaluated under a risk perspective \cite{mehta2018backprop}.

The contributions of this work are:
\begin{itemize}
    \item A Conditional  Generative  Model to correctly predict covariances;
    \item Hybrid approach combining a model-based motion prediction from the Social Force Model(SFM) \cite{helbing1995social} and our implementation of a learning-based Goal Prediction;
    \item Metric to measure the correctness of the uncertainty prediction by counting and averaging the number of times that the prediction lies in the iso-contours of a bi-variate Gaussian.
\end{itemize}

\begin{figure*}[htb]
    \centering
    \includegraphics[width=.80\textwidth]{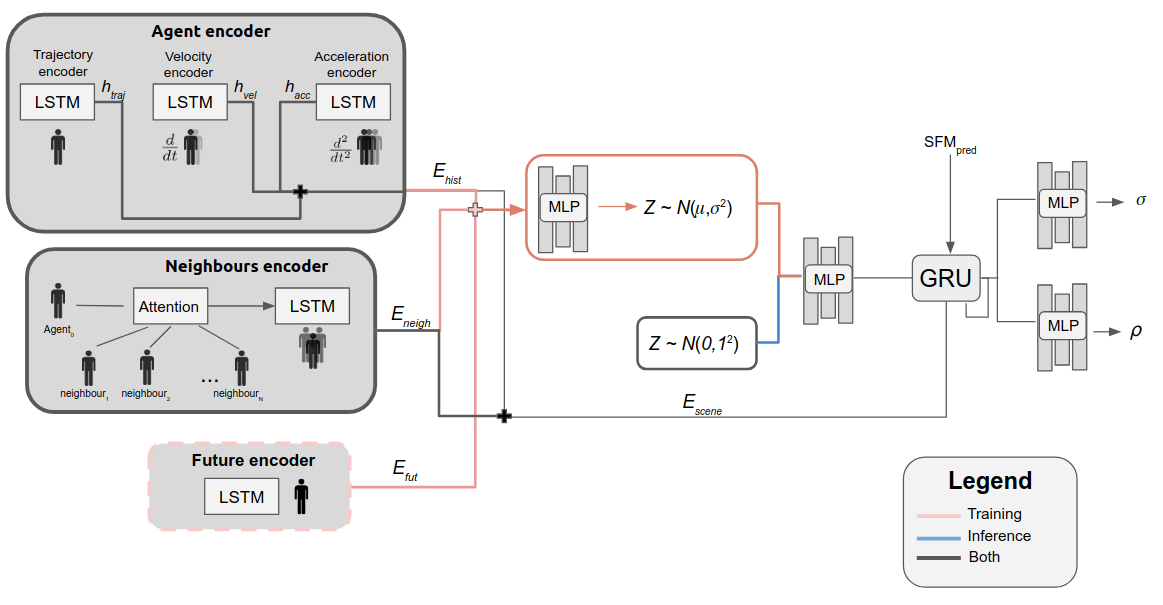}
    \caption{CovarianceNET architecture. CovarianceNET is a part of our proposed hybrid approach for Human Motion Prediction and Uncertainty Estimation. The input consists of the spatial coordinates over the past $T_{obs}$ seconds of all agents in the scene. We use LSTM-based encoder for the agents' history. Neighbours' impact on the predicted motion is encoded by adding an attention module. We use the CVAE latent variable framework for diverse but realistic uncertainty prediction\cite{sohn2015learning}.}
    \label{fig::covNet}
\end{figure*}

\section{Related Work}


Motion prediction has been studied by the robotics community mainly motivated by its direct relation to robot navigation in social environments, embedded at its deepest core.
Examples of robot navigation include Human-aware approaches \cite{sisbot2007human, ferrer2017robot}, considering prediction into planning \cite{ziebart2009planning,fulgenzi2009probabilistic,kuderer2012feature,trautman2015robot} or the effect of planning to prediction \cite{ferrer2014proactive,ferrer2019anticipative,sadigh2016planning}.

Other fields have studied human motion prediction, where multiple previous works study the problem of how to forecast pedestrian trajectories and predict their future behaviors.
Broadly, human motion prediction can be divided in two classes: model-based that creates an empirical model of these transition functions such as the SFM\cite{helbing1995social} and its variants\cite{zanlungo2011social, farina2017walking} or models proposed from the graphics community \cite{van2011,guy2009clearpath}.
And Learning-based approaches \cite{salzmann2020trajectron,alahi2016social,messaoud2019non,giuliari2020transformer,mohamed2020social,yi2016pedestrian} that are becoming the dominant paradigm in human motion prediction, as well as in other topics due to its  unrivaled results. Both of these approaches have the same input and output data and predict the next state of the pedestrians in $dt$ time.

Helbing and Moln\'ar \cite{helbing1995social} have proposed a model based approach for modelling pedestrians' behaviour. Authors have shown that pedestrian motion can be described as a sum of social forces that depend on agents destination point, other pedestrians, static object (borders of buildings, walls, streets, obstacles).

Alahi et al.~\cite{alahi2016social} propose a learning-based method with social pooling, where the Long Short-term Memory(LSTM)\cite{hochreiter1997long} states of neighboring agents were pooled based on their locations in a 2-D grid to form social tensor. 

Messaoud et al.~\cite{messaoud2019non}  apply a multihead attention mechanism to the social tensor to directly relate distant vehicles and extract a context representation.


Recently, generative approaches have emerged as state-of-the-art trajectory forecasting methods due to recent advancements in deep generative models.
They have caused a shift from focusing on predicting the single best trajectory to producing a distribution of potential future trajectories. Most works in this category use a deep recurrent backbone architecture with a latent variable model, such as a Conditional Variational Autoencoder (CVAE) \cite{sohn2015learning} or a Generative Adversarial Network (GAN) \cite{goodfellow2014generative}.

Trajectron++ \cite{salzmann2020trajectron,ivanovic2019trajectron} - a multi-agent behavior prediction model that accounts for the dynamics of the agents, produces predictions possibly conditioned on potential future robot trajectories which can effectively use heterogeneous data about the surrounding environment.

Jein et al. \cite{jain2020discrete} proposes a discrete residual flow to recursively updates the predicted distribution over a pedestrian’s future position in the form of occupancy maps. 

Mahgalam et al.~\cite{mangalam2020not}  propose to address human trajectory prediction by modeling intermediate stochastic goals proposing a socially compliant, endpoint conditioned variational auto-encoder with a novel self-attention based social pooling layer.

In our previous work \cite{postnikov2020hsfm}, we propose a new method for motion prediction - HSFM-$Σ$NN that combines two different approaches: a feed-forward network whose layers are model-based transition functions using the Headed Social Force Model(HSFM) and a Neural Network for covariance prediction.
CovarianceNet is the next step in this work, where we have improved the approach substantially.

Gal et al., \cite{gal2016dropout} proposed a simple yet effective method for probabilistic interpretation of dropout which allows to obtain model uncertainty out of existing deep learning models.

Guo et al. \cite{guo2017calibration} introduced a temperature scaling,  confidence calibration method, that can effectively correct the miscalibration in modern deep neural networks.

\section{Method}

\subsection{Problem formulation}

The position of a generic agent $i$ at time $t$ is represented by $\bm{x}_i^t = (x_i,y_i)_t$, where $(x_i,y_i)_t$ are the coordinates of agents in the global reference system at the instant of time $t$. The agent's trajectory is defined as $X_i^{1:T} = \{ \bm{x}_i^1, ... , \bm{x}_i^T \}$ from timestamp $1$ to $T$.
We aim to generate plausible trajectory distributions for a time-varying number of interacting agents.
Every trajectory is split into observed and future: given certain number $T_{obs}$ of observed positions, we seek a distribution over all agents’ future states for the next $T_{pred}$ time steps which is denoted as $p( X_i^{T_{obs}+1:T_{pred}}| X_i^{1:T_{obs}})$.


Our approach is visualized in Fig. \ref{fig::architecture}.
The first block is the model-based SFM \cite{helbing1995social}, a method for motion prediction the mean positions of the agents' future states.

\begin{equation}
    \textit{sfm}^{T_{obs}+1:T_{pred}}_{1:N} = \textit{SFM}(X^{1:T_{obs}}_{1:N}, \hat{X}^T_{goal_{1:N}})
\end{equation}

At a higher level the SFM needs to infer future possible destinations of pedestrians \cite{ferrer2014}.
In particular, we have customized the learning-based approach by \cite{giuliari2020transformer} and goals are predicted as: 
\begin{equation}
    \hat{X}^T_{goal_{1:N}} = \phi(X^{1:T_{obs}}_{1:N}),
\end{equation}
where $\phi$ - deep multihead-attention based model that trained to predict the position of pedestrian at timestamp $T_{pred}$, $X^{1:T_{obs}}_{1:N}$ - the trajectory (scene history) which contains the states of all $N$ agents at timestamp $t$,  $X^t_{1:N} = \{X^t_1, .., X^t_N\}$, $P_{goal_{1:N}}$ - the position of the pedestrian at timestamp $T_{pred}$, $\textit{sfm}^t_{n}$ - predicted position of $n$ agent at timestamp $t$. 
Here we emphasize that in this work our main goal is to show  correct covariance prediction, thus we are not showing implementation details of Goal Prediction model and SFM.


The CovarianceNet network is one of the sub-blocks of our approach, depicted in Fig. \ref{fig::architecture} and it is shown in detail in Fig.~\ref{fig::covNet}. CovarianceNet is designed to complement the predictions by the SFM  with accurate $x,y$ variances and its correlation coefficient, such that the bi-variate Gaussian constructed from the mean and covariance matrix accurately assesses the potential trajectory distribution of pedestrians, and statistically ensure the properties of a bi-variate Gaussian distribution.

\subsection{Agent encoder} 
For each pedestrian in the scene, the current position and all observed previous positions are known. 
In our work (see Fig.~\ref{fig::covNet}), we are using the {\em Agent} encoder to encode information about agent position and the {\em Neighbours} encoder for encoding information about the influence of the agent and all his neighbors to each other. 
Knowing the history of the agent's movement, one can calculate velocities and accelerations as a additional source of data.



The observed trajectory of each agent $X^{1:T_{obs}}_{1:N}$ is encoded using an LSTM encoder
\begin{equation}
    h_{traj_{i}}^t = \textit{LSTM}(X^{1:T_{obs}}_{1:N}, h_{traj_{i}}^{t-1}, W_{t_{enc}})
    \label{lstm_traj}
\end{equation}

Here, $h_{traj_{i}}^t$ is the hidden state vector of the $i^{th}$ agent at time $t$. All LSTM encoders share the same weights $W_{t_{enc}}$.

Additionally,  we use velocities and accelerations encoded in the same fashion as in (\ref{lstm_traj}) and sum them into a single encoded vector:
\begin{equation}
    h_{acc_{i}}^t = \textit{LSTM}(\frac{\partial }{\partial t}X^{1:T_{obs}}_{1:N}, h_{i_{vel}}^{t-1}, W_{v_{enc}})
\end{equation}
\begin{equation}
    h_{vel_{i}}^t = \textit{LSTM}(\frac{\partial^2 }{\partial t^2}X^{1:T_{obs}}_{1:N}, h_{i_{vel}}^{t-1}, h_{i_{acc}}^{t-1}, W_{a_{enc}})
\end{equation}
\begin{equation}
    E_{hist_i}^t = h_{traj_{i}}^t + h_{vel_{i}}^t + h_{acc_{i}}^t;
\end{equation}
where $h^t_{i_{traj}}$ - encoded trajectory, $h^t_{i_{vel}}$ - encoded velocities, $h^t_{i_{acc}}$ - encoded accelerations and $E^t_{hist}$ - full encoded agent's history.

During the training phase, as it is shown in Fig.\ref{fig::covNet}, to produce a latent distribution additionally leveraged future trajectory encoding $E^t_{fut_i}$ for time indexes ${T_{obs}+1...T_{pred}}$, in the same manner as in Eq.(\ref{lstm_traj}). 

\subsection{Neighbours encoding}
To model neighboring agents’ influence on the modeled agent, all agents' neighbours in the scene are processed with the additive attention module\cite{bahdanau2014neural} and are encoded by the LSTM cell to produce a single neighboring agents’ influence vector $E_{neigh_i}^t$.  
\begin{equation}
    C_i^t = \textit{Attention}(X^{1:T_{obs}}_{neigh_i}, X^{1:T_{obs}}_{i})
\end{equation}
\begin{equation}
    E_{neigh_i}^t = \textit{LSTM}(C_i^t, E_{neigh_i}^{t-1}, W_{neigh})
\end{equation}
\begin{equation}
    E_{scene_i}^t = E_{neigh_i}^t + E_{hist_i}^t
\end{equation}
where $C_i^t$ - context vectors, $E_{neigh_i}^t$ - encoded neighbours influence on the modeled agent, $E_{scene_i}^t$ is a full encoded scene history.

\subsection {Conditional Variational Autoencoder}
The network backbone of our approach (Fig.~\ref{fig::covNet}) is realized as a version of the Conditional Variational  Autoencoder\cite{sohn2015learning} (CVAE). 

The CVAE architecture can be divided in two parts: the prior and the posterior networks. Both prior and posterior distributions are assumed to be Normal distributions. The parameters of the prior are computed by the prior network which only takes the encoded history as input. The posterior parameters are determined from both the encoded history and the target trajectory.
The prior distribution is a Normal distribution, denoted as $p_{\phi_i}(z | X^{1:T_{obs}}_i) = \mathcal{N}(\mu_{prior_i}(x),\sigma^2_{prior_i}(x))$.

\begin{equation}
  p_{\phi_i}(z | X^{1:T_{obs}}_i) = \mathcal{N}(\mu_{prior_i}, \sigma^2_{prior_i})
\end{equation}
\begin{equation}
  \mu_{prior_i} = \textit{MLP}(E_{scene_{i}})
\end{equation}
\begin{equation}
  \sigma_{prior_i} = \textit{MLP}(E_{scene_{i}})
\end{equation}


During training, the latent variable $z$ will be sampled from the posterior distribution.
Specifically, it takes both the encoded past and the encoded future trajectories information $X_i$ passed through an MLP to obtain a latent mean $\mu_{latent}$ and a latent sigma $\sigma_{latent}$ to output a latent distribution $q_{\phi_i}(z | X^{T_{obs}+1:T_{pred}}_i, X^{1:T_{obs}}_i)$.



\begin{equation}
  q_{\phi_i}(z | X^{T_{obs}+1:T_{pred}}_i, X^{1:T_{obs}}_i) = \mathcal{N}(\mu_{latent_i}, \sigma^2_{latent_i})
\end{equation}
\begin{equation}
  \mu_{latent_i} = \textit{MLP}(E_{scene_{i}}, E_{fut_{i}})
\end{equation}
\begin{equation}
  \sigma_{latent{i}} = \textit{MLP}(E_{scene_{i}}, E_{fut_{i}})
\end{equation}

In our work, we use two fully connected layers with dropout and sigmoid activation function between them as base multi layer perceptron (MLP).

\subsection{Decoder}
The decoder models the probability of a target trajectory $X^{T_{obs}+1..T_{pred}}$ given
the latent variable $z$ sampled from the latent distribution, encoded node history $E_{scene_{i}}^t$ and SFM trajectory predictions $\textit{sfm}^t_i$.
 
We make an assumption that the target distribution $p(\bm{x}^{t} | X^{1:T_{obs}})$ is a bi-variate Gaussian:

\begin{equation}
        p_i(\bm{x}^{t}_i|X^{1:T_{obs}}_i) = N(\mu^t_i,\Sigma^t_i)
\end{equation}
\begin{equation}
        \mu^{t}_i = \begin{bmatrix}
                \mu_{x_i}^{t}\\
                \mu_{y_i}^{t}\\
                \end{bmatrix};       
        \Sigma^t_i = \begin{bmatrix}
                    \sigma_{x_i}^{t^2}  & \rho\sigma_{x_i}^t\sigma_{y_i}^t \\
                    \rho\sigma_{x_i}^t \sigma_{y_i}^t   &   \sigma_{y_i}^{t^2}\\
                \end{bmatrix}.
\end{equation}

For each timestamp of prediction, we use encoded information about current agent state, passed through the MLP as an input to the GRU layer with $E_{scene}$ as the initial state of the GRU cell, witch recurrently outputs the new hidden states for each agent.
\begin{equation}
    h^t_{gru_i} = \textit{GRU}([\textit{sfm}^{t-1}_i, z], h^{t-1}_{gru_i})
\end{equation}

GRU hidden states are then used to predict the parameters of a bi-variate Gaussian distribution $\mathcal{N}(\mu, \sigma, \rho)$  standard deviations $\sigma$ and correlation coefficient $\rho$, while the mean values $\mu$ are taken from the SFM prediction $\mu_i = \textit{sfm}^t_i$.

\begin{equation}
  \rho^t_i = \textit{MLP}(h^t_{gru_i})
\end{equation}
\begin{equation}
  \sigma^t_i = \textit{MLP}(h^t_{gru_i})
\end{equation}

The entire model is trained by maximising the sequential evidence lower-bound (ELBO):
\begin{equation}
    L_{nll}(W) = -\sum^{N}_{i=1}\sum^{T_{pred}}_{t=T_{obs}+1}log(p^t_{i}(\bm{x}^t_i| \mu^t_i, \sigma^t_i,\rho^t_i))
    \label{eq::NLL}
\end{equation}

\begin{adjustwidth}{-0.25cm}{}
\begin{equation}
    L^n_{kl} =\sum^{N}_{i=1}D_{kl}(q_{\phi_i}(z|X^{T_{obs}+1:T_{pred}}_i, X^{1:T_{obs}}_i)||N(0,1))
\end{equation}
\end{adjustwidth}
\begin{equation}
    \textit{Loss} = \alpha  L^n_{nll}(W) + L^n_{kl}
\end{equation}

It is important to mention that in our case, when we utilize Goal Predictor and SFM\cite{helbing1995social} as model for mean poses prediction we do not optimize $\mu^t_i$ parameter from (\ref{eq::NLL}).



\begin{figure*}[ht!]
    \centering
    \includegraphics[width=8cm]{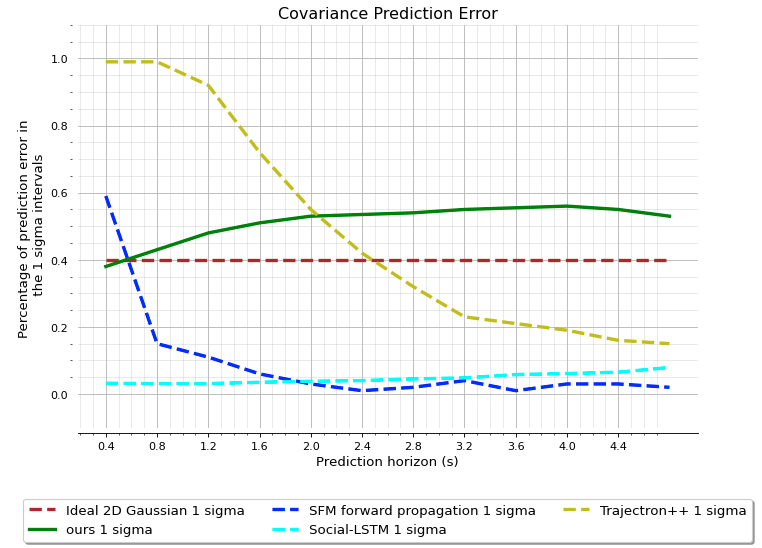}
    \includegraphics[width=8cm]{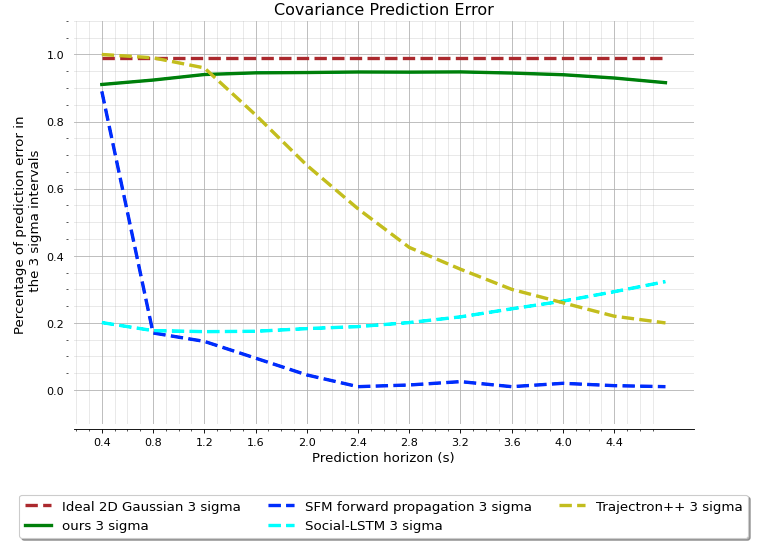}
    \caption{Evaluation of $\textit{PPEI}$. {\em Left:} results for $1\sigma$. {\em Right:} results for $3\sigma$.
    }
    \label{fig:sigmaErr}
\end{figure*}

\section{ Evaluation}

In our work, we use the combination of goal predictor and SFM\cite{helbing1995social} modules as a base predictor for mean future poses of agents and combine it with the CovarianceNet in order to predict accurate uncertainties. We compare results of our method with other three popular approaches for evaluating uncertainty on publicly-available ETH\cite{pellegrini2009you} pedestrian dataset.
It consists of real world human trajectories with rich multi-human interaction scenarios.
This dataset is a standard benchmark in the field as it contain challenging behaviors such as couples walking together, groups crossing each other, and groups forming and dispersing.  
Leave-one-out strategy was used for evaluation, where the model is trained on four datasets and evaluated on the held-out fifth. 
An observation length of 8 timesteps (3.2s) and a prediction horizon of 12 timesteps (4.8s) is used for evaluation. In this section we will compare results of our method with the following methods:

\subsection{Evaluated Methods}

\subsubsection{\em \textit{Covariance Forward-Propagation (FP)}}

In our work we have used SFM for prediction forces acting on all agents in scene and Human locomotion model which integrates predicted forces to the state variables of a pedestrians 2D poses at timestamp $t+1$, in the following generalized way:
\begin{equation}
    x_{t+1}= T(x_{t})
    \label{eq_trans}
\end{equation}

The transition function $T()$ defined by the SFM (\ref{eq_trans}) is a non-linear differentiable function (by construction). The simplest method for covariance estimation is using the first-order Taylor expansion:
\begin{equation}
    x_{t+1}= T(\mu_{t}) + G_{t}(x_{t}-\mu_{t})
\end{equation}
where $\mu_{t}$ is the current state estimate and $G_{t}$ is the Jacobian matrix $\partial T / \partial x$ evaluated at $\mu_t$. From here, we apply {\em Covariance Propagation} of a Gaussian random variable ($x_{t} \sim \mathcal{N}( \mu_{t},\Sigma_{t})$) over a linear function:
\begin{equation}
    x_{t+1}  \sim \mathcal{N}\big(T(\mu_{t}),  G_{t} \cdot \Sigma_{t}\cdot  G_{t}^{\top}\big)
\label{eq_fp}
\end{equation}

\subsubsection{\textit{Social-LSTM} \cite{alahi2016social}}
One of the most influential works in direction of Human Motion Distribution Prediction based on Neural  Networks, where each agent is modeled with an LSTM and nearby agents’ hidden states are pooled at each timestep using a proposed social pooling operation.

\subsubsection{\textit{Trajectron++} \cite{salzmann2020trajectron}}
One of the latest works in this direction is named \textit{Trajectron++}\cite{salzmann2020trajectron} has four configurations of predictions, of which we used Most Likely, which gives the best results of ADE and FDE. Trajectron++ calculates the result of predictions as a Gaussian Mixture Model (GMM) witch contains 25 Gaussian distributions.


\subsection{Motion Prediction Evaluation}
We use Euclidean distance errors, Average Displacement Error (ADE) and Final Displacement Error (FDE) to evaluate the accuracy of then mentioned approaches in a motion prediction task. Metrics are formulated as: 

\begin{equation}
    ADE^t = \frac{\sum_{j=1}^N ||\bm{x}^t_j -\mu^t_j||_2}{N}
\end{equation}
\begin{equation}
    FDE = \frac{\sum_{j=1}^N{ ||\bm{x}^{T_{pred}}_j -\mu^{T_{pred}}_j||_2}}{N}
\end{equation}
where N - number of processed pedestrians, $\bm{x}_j^t$ - ground truth position of $\textit{j}^{th}$ pedestrian at timestamp \textit{t}, $\textit{T}_{pred}$ - prediction horizon, $\mu$ - predicted mean position.

The results in Table \ref{table:1} shows that modern deep approaches are superior to model-based approaches in terms of ADE and FDE. The Trajectron's++\cite{salzmann2020trajectron} approach takes into consideration a multimodal distribution, which results in the best Euclidean errors. 
Still, the contributions of this paper are on the covariance prediction, and combination of SFM and Goal Predictor could be substituted by any modern prediction method.


\subsection{Covariance Estimation Evaluations}\label{sec::NN}

We propose to evaluate the accuracy of the covariance prediction methods by the following metric, the Part of Prediction Errors Inside:
\begin{equation}
    \textit{PPEI}_\alpha = E\{\bm{1}( ||\bm{x} -\mu||_{\Sigma} < \alpha) \}, \quad \alpha = \{1,3\}.
\end{equation}
where $\bm{1}()$ is the indicator function and we simply average the number of times that our prediction lies
inside the  $1\sigma$ and $3\sigma$ ellipsoids from the ground truth position by using the Mahalanobis distance.


\begin{figure}[h]
    \centering
    \includegraphics[width=8.5cm]{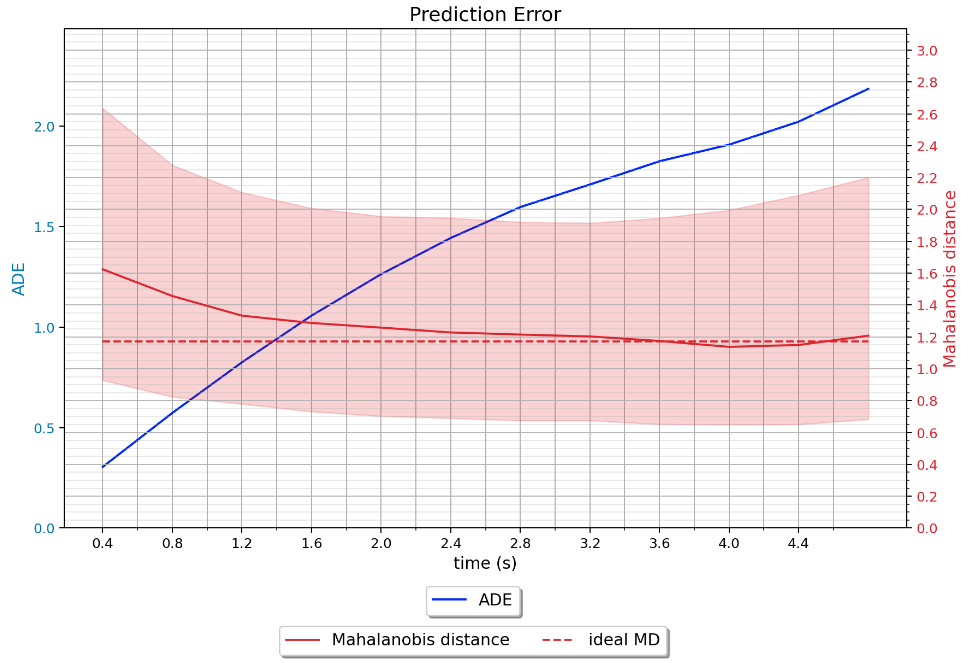}
    \caption{Mahalanobis  error (in red)  distances and ADE(in blue)  for  our proposed method.  Mahalanobis  solid  line  is  median values, and colored intervals are .25 and .75 percentiles.}
    \label{fig:ADEMD_2021}
\end{figure}


\begin{table*}[h!]
\caption{Comparison of our method against previously published methods on the ETH dataset \cite{pellegrini2009you}. Both
ADE and FDE are reported in meters and presented here for reference(*is not valid since ground truth last pedestrian pose is used as input). Our main contribution is accurately predicted uncertainties, which can be measured as part of errors inside $1\sigma$ $3\sigma$ interval and its deviation from the ideal 2D Gaussian.}
\label{table:1}
\begin{center}
        \begin{tabular}{||c c c c c c||} 
             \hline
             Method & ADE & FDE 
             &   \vtop{\hbox{\strut \% errors inside $1\sigma$ }
                  \hbox{\strut   ($\Delta$ from expected)}}  
             &   \vtop{\hbox{\strut \% errors inside $3\sigma$}
                  \hbox{\strut   ($\Delta$ from expected) }} & median Mahalanobis Distance\\
            
             \hline\hline 
             Social-LSTM\cite{alahi2016social}              & 1.09     & 2.35  &  4.7$\pm$11(35) & 22 $\pm$11(76.7) & 3.4$\pm$5.66 \\  
             Trajectron++ (Distribution) \cite{salzmann2020trajectron}    & \textbf{0.71}    & \textbf{1.66}  & \textbf{50}$\pm$32(\textbf{10.3}) & 56$\pm$30(42.4) & --\\
             SFM + FP & 0.98* & 0.2*  & 8.2$\pm$15(31.5) & 11.6$\pm$23(87.2) & -- \\
             \hline\hline
             CovarianceNet & 1.39 & 2.18  & 51.2\textbf{$\pm$0.3}(11.5) & \textbf{93.7$\pm$0.01(5)} & 1.21$\pm$2.16 \\
             \hline
        \end{tabular}
    \end{center}
\end{table*}

\begin{figure*}[h]
     \centering
    \includegraphics[width=10.7cm]{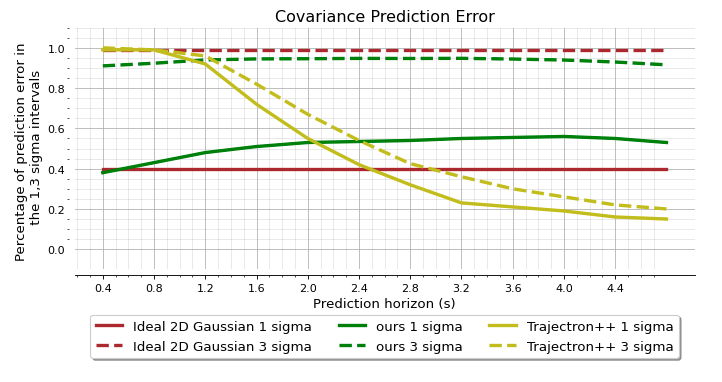}
\caption{\textit{PPEI} Evaluation of CovarianceNet method and Trajectron++(Distribution configuration) qualities of predicted uncertainties peaking Trajectron++ Gaussian with the highest probability (out of GMM 25 Gaussians component).}
    \label{fig::trajectron}
\end{figure*}

Figure \ref{fig:sigmaErr} shows the $\textit{PPEI}_1$ and $\textit{PPEI}_3$ for each of the methods described above.
 Our proposed method achieves a superior performance (Fig \ref{fig:sigmaErr}) compared to the previously published methods in terms of performance of predicted uncertainties.

It can be seen in Fig.~\ref{fig:sigmaErr} and in Table \ref{table:1} that modern deep approaches output accurate ADE, FDE but become overconfident when predicting covariances of distributions, while proposed method with sequentially trained Goal Predictor and CovarianceNet models  are capable to predict statistically correct covariances. Despite the fact that Trajectron's++ average  $PPEI_1$ over all horizon time is close to ideal, it clearly can be observed a high deviation, ranging from a high $PPEI_1$ (under confident) at initial horizon times to low values at the final horizon time (over confident), which is shown with high variance, $PPEI_1$=$50$ \textbf{$\pm32$} at Table \ref{table:1}, while our proposed method produces more consistent predictions $PPEI_1$=$51.2$ \textbf{$\pm0.3$}.

The forward propagation (FP) method collapses and provides poor results due to vanishing gradients over multiple iterations.
Trajectron shows a decreasing value of $\textit{PPEI}$, becoming clearly overconfident.
Social-LSTM also provides a overconfident covariance estimations, with mean $1\sigma$ PPEI=4.7\% and $3\sigma$ PPEI=22\%.


In Fig.~\ref{fig:ADEMD_2021} we show the median, 25 and 75 percentiles of the Mahalanobis distances for the predicted distributions of CovarianceNet and the Average Displacement Error(ADE) over prediction horizon time for predicted mean trajectories, with dashed red line we show median Mahalanobis Distance (MD) for ideal bi-variate Gaussian.
Our projected median MD for the entire prediction horizon is 1.21, which is close to the bi-variate Normal distribution, with a median MD of 1.17.


We use Trajectron++\cite{salzmann2020trajectron} in Distribution configuration to estimate probability distribution. We used only one Gaussian distribution from GMM for the metric at each prediction step, which had the highest probability. An example of such a prediction and that interpretations can be seen in Fig.~\ref{fig:example}. This evaluation interpretations have similar to previous interpretations results, with overconfident distribution for large prediction horizons and under confident for smaller time horizons.

    

Fig \ref{fig::trajectron} shows that individual Gaussians, composing GMM, is far from ideal, which potentially can lead to unexpected results when a person is outside the probabilistic estimate of the pedestrian position.

\section{Conclusions}
We have presented a new hybrid method, CovarianceNet, that combines model-based  Human Motion Prediction with a neural network approach for covariance  prediction. Our approach brings an efficient calculation of motion prediction, by using the SFM recursively and calculates the uncertainty associated with this prediction by using a conditional deep generative model CVAE with Gaussian latent variables.

We have evaluated our results in the ETH dataset and compared with state-of-the art approaches.
While Social-LSTM and Tajectron methods show better accuracy when evaluating the prediction error, they are overconfident on their predicted distributions, according to our proposed metric to measure uncertainty prediction. Our method, CovarianceNet, is able to predict correctly its uncertainty, thereby, our approach is better suited for applications requiring accurate prediction of distributions.

\bibliographystyle{ieeetr}
\balance
\bibliography{root.bib}

\begin{thebibliography}{10}

\bibitem{pellegrini2009you}
S.~Pellegrini, A.~Ess, K.~Schindler, and L.~Van~Gool, ``You'll never walk
  alone: Modeling social behavior for multi-target tracking,'' in {\em 2009
  IEEE 12th International Conference on Computer Vision}, pp.~261--268, IEEE,
  2009.

\bibitem{leal2014learning}
L.~Leal-Taix{\'e}, M.~Fenzi, A.~Kuznetsova, B.~Rosenhahn, and S.~Savarese,
  ``Learning an image-based motion context for multiple people tracking,'' in
  {\em Proceedings of the IEEE Conference on Computer Vision and Pattern
  Recognition}, pp.~3542--3549, 2014.

\bibitem{caesar2020nuscenes}
H.~Caesar, V.~Bankiti, A.~H. Lang, S.~Vora, V.~E. Liong, Q.~Xu, A.~Krishnan,
  Y.~Pan, G.~Baldan, and O.~Beijbom, ``nuscenes: A multimodal dataset for
  autonomous driving,'' in {\em Proceedings of the IEEE/CVF Conference on
  Computer Vision and Pattern Recognition}, pp.~11621--11631, 2020.

\bibitem{salzmann2020trajectron}
T.~Salzmann, B.~Ivanovic, P.~Chakravarty, and M.~Pavone, ``Trajectron++:
  Multi-agent generative trajectory forecasting with heterogeneous data for
  control,'' {\em arXiv preprint arXiv:2001.03093}, 2020.

\bibitem{mehta2018backprop}
D.~Mehta, G.~Ferrer, and E.~Olson, ``Backprop-mpdm: Faster risk-aware policy
  evaluation through efficient gradient optimization,'' in {\em 2018 IEEE
  International Conference on Robotics and Automation (ICRA)}, pp.~1740--1746,
  IEEE, 2018.

\bibitem{helbing1995social}
D.~Helbing and P.~Moln\'ar, ``Social force model for pedestrian dynamics,''
  {\em Physical review E}, vol.~51, no.~5, p.~4282, 1995.

\bibitem{giuliari2020transformer}
F.~Giuliari, I.~Hasan, M.~Cristani, and F.~Galasso, ``Transformer networks for
  trajectory forecasting,'' {\em arXiv preprint arXiv:2003.08111}, 2020.

\bibitem{sohn2015learning}
K.~Sohn, H.~Lee, and X.~Yan, ``Learning structured output representation using
  deep conditional generative models,'' in {\em Advances in neural information
  processing systems}, pp.~3483--3491, 2015.

\bibitem{sisbot2007human}
E.~A. Sisbot, L.~F. Marin-Urias, R.~Alami, and T.~Simeon, ``A human aware
  mobile robot motion planner,'' {\em IEEE Transactions on Robotics}, vol.~23,
  no.~5, pp.~874--883, 2007.

\bibitem{ferrer2017robot}
G.~Ferrer, A.~G. Zulueta, F.~H. Cotarelo, and A.~Sanfeliu, ``Robot social-aware
  navigation framework to accompany people walking side-by-side,'' {\em
  Autonomous robots}, vol.~41, no.~4, pp.~775--793, 2017.

\bibitem{ziebart2009planning}
B.~D. Ziebart, N.~Ratliff, G.~Gallagher, C.~Mertz, K.~Peterson, J.~A. Bagnell,
  M.~Hebert, A.~K. Dey, and S.~Srinivasa, ``Planning-based prediction for
  pedestrians,'' in {\em IEEE/RSJ International Conference on Intelligent
  Robots and Systems}, pp.~3931--3936, IEEE, 2009.

\bibitem{fulgenzi2009probabilistic}
C.~Fulgenzi, A.~Spalanzani, and C.~Laugier, ``Probabilistic motion planning
  among moving obstacles following typical motion patterns,'' in {\em IEEE/RSJ
  International Conference on Intelligent Robots and Systems}, pp.~4027--4033,
  IEEE, 2009.

\bibitem{kuderer2012feature}
M.~Kuderer, H.~Kretzschmar, C.~Sprunk, and W.~Burgard, ``Feature-based
  prediction of trajectories for socially compliant navigation.,'' in {\em
  Robotics: science and systems}, 2012.

\bibitem{trautman2015robot}
P.~Trautman, J.~Ma, R.~M. Murray, and A.~Krause, ``Robot navigation in dense
  human crowds: Statistical models and experimental studies of human--robot
  cooperation,'' {\em The International Journal of Robotics Research}, vol.~34,
  no.~3, pp.~335--356, 2015.

\bibitem{ferrer2014proactive}
G.~Ferrer and A.~Sanfeliu, ``Proactive kinodynamic planning using the extended
  social force model and human motion prediction in urban environments,'' in
  {\em 2014 IEEE/RSJ International Conference on Intelligent Robots and
  Systems}, pp.~1730--1735, IEEE, 2014.

\bibitem{ferrer2019anticipative}
G.~Ferrer and A.~Sanfeliu, ``Anticipative kinodynamic planning: multi-objective
  robot navigation in urban and dynamic environments,'' {\em Autonomous
  Robots}, vol.~43, no.~6, pp.~1473--1488, 2019.

\bibitem{sadigh2016planning}
D.~Sadigh, S.~Sastry, S.~A. Seshia, and A.~D. Dragan, ``Planning for autonomous
  cars that leverage effects on human actions.,'' in {\em Robotics: Science and
  Systems}, vol.~2, Ann Arbor, MI, USA, 2016.

\bibitem{zanlungo2011social}
F.~Zanlungo, T.~Ikeda, and T.~Kanda, ``Social force model with explicit
  collision prediction,'' {\em EPL (Europhysics Letters)}, vol.~93, no.~6,
  p.~68005, 2011.

\bibitem{farina2017walking}
F.~Farina, D.~Fontanelli, A.~Garulli, A.~Giannitrapani, and D.~Prattichizzo,
  ``Walking ahead: The headed social force model,'' {\em PloS one}, vol.~12,
  no.~1, p.~e0169734, 2017.

\bibitem{van2011}
J.~Van Den~Berg, S.~J. Guy, M.~Lin, and D.~Manocha, ``Reciprocal n-body
  collision avoidance,'' in {\em Robotics research}, pp.~3--19, Springer, 2011.

\bibitem{guy2009clearpath}
S.~J. Guy, J.~Chhugani, C.~Kim, N.~Satish, M.~Lin, D.~Manocha, and P.~Dubey,
  ``Clearpath: highly parallel collision avoidance for multi-agent
  simulation,'' in {\em Proceedings of the ACM SIGGRAPH/Eurographics Symposium
  on Computer Animation}, pp.~177--187, 2009.

\bibitem{alahi2016social}
A.~Alahi, K.~Goel, V.~Ramanathan, A.~Robicquet, L.~Fei-Fei, and S.~Savarese,
  ``Social {LSTM}: Human trajectory prediction in crowded spaces,'' in {\em
  Proceedings of the IEEE conference on computer vision and pattern
  recognition}, pp.~961--971, 2016.

\bibitem{messaoud2019non}
K.~Messaoud, I.~Yahiaoui, A.~Verroust-Blondet, and F.~Nashashibi, ``Non-local
  social pooling for vehicle trajectory prediction,'' in {\em 2019 IEEE
  Intelligent Vehicles Symposium (IV)}, pp.~975--980, IEEE, 2019.

\bibitem{mohamed2020social}
A.~Mohamed, K.~Qian, M.~Elhoseiny, and C.~Claudel, ``Social-stgcnn: A social
  spatio-temporal graph convolutional neural network for human trajectory
  prediction,'' in {\em Proceedings of the IEEE/CVF Conference on Computer
  Vision and Pattern Recognition}, pp.~14424--14432, 2020.

\bibitem{yi2016pedestrian}
S.~Yi, H.~Li, and X.~Wang, ``Pedestrian behavior understanding and prediction
  with deep neural networks,'' in {\em European Conference on Computer Vision},
  pp.~263--279, Springer, 2016.

\bibitem{hochreiter1997long}
S.~Hochreiter and J.~Schmidhuber, ``Long short-term memory,'' {\em Neural
  computation}, vol.~9, no.~8, pp.~1735--1780, 1997.

\bibitem{goodfellow2014generative}
I.~Goodfellow, J.~Pouget-Abadie, M.~Mirza, B.~Xu, D.~Warde-Farley, S.~Ozair,
  A.~Courville, and Y.~Bengio, ``Generative adversarial nets,'' in {\em
  Advances in neural information processing systems}, pp.~2672--2680, 2014.

\bibitem{ivanovic2019trajectron}
B.~Ivanovic and M.~Pavone, ``The trajectron: Probabilistic multi-agent
  trajectory modeling with dynamic spatiotemporal graphs,'' in {\em Proceedings
  of the IEEE International Conference on Computer Vision}, pp.~2375--2384,
  2019.

\bibitem{jain2020discrete}
A.~Jain, S.~Casas, R.~Liao, Y.~Xiong, S.~Feng, S.~Segal, and R.~Urtasun,
  ``Discrete residual flow for probabilistic pedestrian behavior prediction,''
  in {\em Conference on Robot Learning}, pp.~407--419, PMLR, 2020.

\bibitem{mangalam2020not}
K.~Mangalam, H.~Girase, S.~Agarwal, K.-H. Lee, E.~Adeli, J.~Malik, and
  A.~Gaidon, ``It is not the journey but the destination: Endpoint conditioned
  trajectory prediction,'' {\em arXiv preprint arXiv:2004.02025}, 2020.

\bibitem{postnikov2020hsfm}
A.~Postnikov, A.~Gamayunov, and G.~Ferrer, ``{HSFM-SigmaNN}: Combining a
  feedforward motion prediction network and covariance prediction,'' {\em arXiv
  preprint arXiv:2009.04299}, 2020.

\bibitem{gal2016dropout}
Y.~Gal and Z.~Ghahramani, ``Dropout as a bayesian approximation: Representing
  model uncertainty in deep learning,'' in {\em international conference on
  machine learning}, pp.~1050--1059, 2016.

\bibitem{guo2017calibration}
C.~Guo, G.~Pleiss, Y.~Sun, and K.~Q. Weinberger, ``On calibration of modern
  neural networks,'' in {\em International Conference on Machine Learning},
  pp.~1321--1330, PMLR, 2017.

\bibitem{ferrer2014}
G.~Ferrer and A.~Sanfeliu, ``Bayesian human motion intentionality prediction in
  urban environments,'' {\em Pattern Recognition Letters}, vol.~44,
  pp.~134--140, 2014.

\bibitem{bahdanau2014neural}
D.~Bahdanau, K.~Cho, and Y.~Bengio, ``Neural machine translation by jointly
  learning to align and translate,'' {\em arXiv preprint arXiv:1409.0473},
  2014.

\end{thebibliography}

\end{document}